# Separating Capability from Permission: A Governance Framework for Agentic AI Autonomy Levels

**Haining Zheng[1], Qian Dong[1], Rodolfo K. Depena[1], Jonathan D. Bhatia[1], Feng Xiao[2], Peng Xu[1]**

[1]ExxonMobil Technology and Engineering Company

[2]ExxonMobil Global Operations Company

**Abstract**

As AI systems increasingly exhibit agentic behavior, discussions of autonomy often conflate what systems are technically capable of doing with what they should be permitted to do in practice. This paper introduces a governance framework that explicitly separates Allowed Autonomy Levels (AAL)—which define the degree of autonomy an AI agent is authorized to exercise given risk, oversight, and accountability considerations—from Autonomous Capability Levels (ACL), which characterize an agent's inherent technical abilities. We present a structured set of autonomy levels spanning reactive execution, decision support, supervised action, goal-directed autonomy, and delegated operational authority, and describe how control, reversibility, and accountability change as autonomy increases. To operationalize this framework, we propose a risk-aware decision process for assigning allowed autonomy, analyze how risk and accountability evolve across autonomy levels, and demonstrate its application through a deployed enterprise data engineering agent, illustrating how a system assessed at a high capability level can be deliberately constrained to a lower allowed autonomy based on risk, reversibility, and organizational readiness. By distinguishing authorization from capability, this work provides practical guidance for the design, deployment, and governance of Agentic AI systems.

## Introduction

Artificial intelligence (AI) technologies and applications are being rapidly adopted across society, ranging from consumer-facing products to large-scale enterprise systems, largely driven by advances in Large Language Models (LLMs). Despite this rapid adoption, it has been widely recognized that the development of Responsible AI frameworks, particularly in corporate enterprises, has not kept pace with advancements in AI technologies (AI Index Steering Committee 2026). This misalignment has become increasingly pronounced with the recent acceleration in the development and deployment of Agentic AI.

A critical requirement for the responsible design, governance, and deployment of Agentic AI systems is the explicit definition of autonomy levels. Without clear and shared autonomy definitions, organizations struggle to reason consistently about oversight requirements, acceptable risk, and accountability boundaries. However, existing work on autonomy classification for AI systems remains limited, in part because such distinctions were not practically necessary until recently. Some approaches define autonomy levels by drawing analogies to autonomous driving systems (SAE 2021). Other perspectives focus on characterizing the role of humans in the decision-making loop (Feng, McDonald, and Zhang 2025), while additional efforts emphasize legal or liability-based considerations (Soder et al. 2025). While valuable, these approaches often conflate an agent's technical capabilities with the degree of authority it is permitted to exercise in practice, or they focus on a single dimension of autonomy without providing an operational decision framework.

In this paper, we propose a governance framework for defining autonomy levels that is explicitly designed to integrate with our enterprise Responsible AI practices in an actionable and operational manner. Building on prior work and informed by analogies to human organizational and decision-making systems, we argue that a clear separation between an AI agent's inherent technical capabilities and the degree of autonomy it is authorized to exercise enables more transparent, accountable, and risk-aware governance. We distinguish between Allowed Autonomy Levels (AAL), which define how much autonomy an agent is permitted to exercise given governance, oversight, and accountability constraints and Autonomous Capability Levels (ACL), which capture what an agent is technically capable of doing.

This separation provides a practical basis for consistent review, approval, and deployment of Agentic AI systems across heterogeneous enterprise environments. By decoupling capability assessment from autonomy authorization, the proposed framework supports scalable governance while accommodating rapid advances in Agentic AI technologies. This framework is intended to help organizations move faster and safer by providing an intuitive and practical governance approach that enables agents to be deployed at appropriate autonomy levels.

## Basic Concepts and Design Principles

The autonomy framework proposed in this paper is rooted in our enterprise Responsible AI Framework, which is developed on the foundation of the NIST AI Risk Management Framework (NIST 2023) and complies with regulations such as the EU AI Act (EU 2024). It has been deployed for two years with an established intake process, risk questionnaires, trained AI risk advisors, and a governance dashboard used to track AI projects. While the system has been working well for traditional machine learning approaches and large foundation models based GenAI applications, we find that it needs to be updated to accommodate recent advances in Agentic AI.

In this work, an **AI agent** is defined as an autonomous or semi-autonomous computer system that perceives its environment, reasons, plans, and takes actions to achieve goals, and may improve through learning. Our definition is based on widely accepted work by (Brustoloni 1991) and (Russell and Norvig 2020). **Agentic AI** refers to an application in which one or more AI agents are composed and orchestrated to perform complex tasks or workflows through the coordination of data, tools, and other systems, operating at varying levels of autonomy within defined constraints and oversight mechanisms. Such systems may consist of multiple interacting agents, orchestration logic, external tools, and supporting infrastructure, collectively producing autonomous behavior that cannot be attributed to any single component in isolation.

The autonomy framework proposed in this paper is guided by three design principles: **rigor**, **stability**, and **usability**. These principles reflect both conceptual considerations about how autonomy should be reasoned about and practical constraints arising from enterprise Responsible AI governance. Together, they shape how autonomy levels are defined, interpreted, and applied across a wide range of Agentic AI systems.

Rigor is required to ensure that autonomy assessments are meaningful, consistent, and defensible. Autonomy levels must be defined using clear criteria that allow different reviewers, teams, and organizations to arrive at consistent conclusions when evaluating similar systems. This objective is particularly important because autonomy assessments may inform deployment approvals, governance controls, and accountability assignments.

Stability is critical because AI technologies evolve rapidly, while governance frameworks must remain applicable over longer time horizons. Autonomy levels that depend on specific model capabilities, benchmarks, or architectural details risk becoming obsolete or misleading as new techniques emerge. To remain stable, autonomy definitions should abstract away from implementation details and focus instead on externally observable behaviors and decision-making authority.

Usability is essential for practical adoption. In large organizations, autonomy assessments are often performed by multidisciplinary teams that include not only AI researchers and engineers, but also legal, risk, and governance professionals. A framework that is overly complex or difficult to interpret risks being inconsistently applied or ignored altogether. For this reason, the number of autonomy levels should be limited, their meanings clearly articulated, and their progression intuitive.

## Autonomy Level Decision Framework

A common approach to defining autonomy levels for AI systems is to draw analogies to human roles or to existing taxonomies from other domains, such as the Society of Automotive Engineers (SAE) levels for autonomous driving systems (SAE 2021). While such approaches are intuitive, they often implicitly combine technical capability and permitted autonomy into a single progression. In practice, this coupling can be problematic for Agentic AI systems, particularly in enterprise environments where authorization, accountability, and oversight are strongly context-dependent.

Encouragingly, some prior work has explicitly centered the role of humans in agentic systems. The framework proposed by (Feng, McDonald, and Zhang 2025) exemplifies this approach by defining five levels of user involvement: L1, where users act as operators and manage planning, invocation, or approval before actions; L2, where users serve as collaborators and control can transfer between users and agents; L3, where users function as consultants while agents plan and execute most tasks; L4, where users act primarily as approvers and are engaged only in high-consequence situations; and L5, where users are observers and agents plan and execute tasks with minimal human involvement aside from extreme interventions such as emergency shutdowns. This framework clearly articulates human roles across varying degrees of autonomy and is applicable to a wide range of settings. However, it does not provide an explicit decision framework for assessing an agent's inherent capabilities or for systematically reasoning about the potential real-world consequences associated with different autonomy assignments.

A useful parallel can be found in human organizations, where capability and authority are routinely decoupled. An individual may be fully capable of independently performing a task, yet still be required to seek approval before doing so, or may be authorized to act only within a limited scope. The autonomy framework proposed in this paper adopts a similar perspective for Agentic AI systems.

Accordingly, we distinguish between two dimensions of autonomy. Allowed Autonomy Levels (AAL) describe the degree of autonomy that an AI agent or agentic system is permitted to exercise in deployment, taking into account governance requirements, oversight mechanisms, accountability structures, and acceptable risk thresholds. Autonomous Capability Levels (ACL) describe what an AI agent or agentic system is technically capable of doing, independent of organizational policy or authorization, based on observable behaviors such as perception, reasoning, planning, tool use, and coordination.

Figure 1: Flowchart to determine autonomy level.

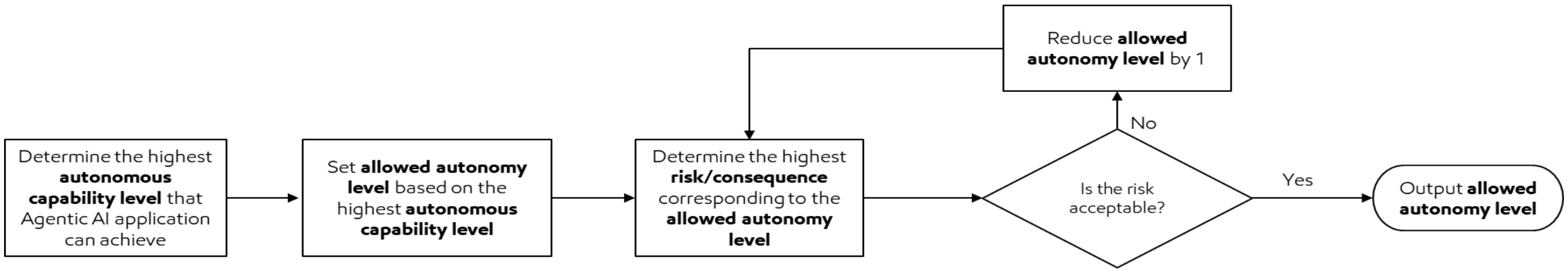

This separation allows decisions about autonomy to be made in a more structured and transparent manner. Capability assessments establish an upper bound on what a system could do, while authorization decisions determine what it should be allowed to do in a given context. Importantly, higher capability does not automatically imply higher allowed autonomy. A highly capable system may still be deployed with limited autonomy, while a less capable system may be granted greater freedom within a tightly constrained scope.

Figure 1 illustrates the proposed autonomy level decision framework. The first step is to assess the Autonomous Capability Level of the system based on its technical characteristics. This assessment constrains the set of Allowed Autonomy Levels that are feasible but does not determine the final authorization decision. Governance considerations—including potential impact, reversibility of actions, accountability structures, and mitigation mechanisms—are then used to select an appropriate Allowed Autonomy Level for deployment.

Note that a key decision step is "Is the risk acceptable?". We leverage our existing risk assessment questionaries and decision process under the Responsible AI Framework. Readers may leverage their existing risk assessment framework or establish such a practice if the framework does not exist.

The decision process terminates at A1, which represents the lowest allowed autonomy level. For most applications, risks at A1 are acceptable given full human control. In rare cases where risk remains unacceptable even at A1, the application should not be deployed in its current form and requires redesign.

This two-step decision process is particularly important for Agentic AI systems, where autonomous behavior may emerge from interactions between multiple components rather than from any single agent. By explicitly separating capability assessment from autonomy authorization, the framework supports stable and reusable governance decisions across diverse systems and deployment scenarios. It also enables incremental and reversible autonomy progression, allowing organizations to increase or decrease permitted autonomy without reinterpreting underlying capabilities. The following sections describe the Allowed Autonomy Levels and Autonomous Capability Levels in detail. Together, they operationalize the decision framework introduced here and provide a structured basis for governing Agentic AI systems within enterprise responsible AI programs.

## Allowed Autonomy Level

Allowed Autonomy Level (AAL) concerns the level of accountability that an organization is willing to assume for an Agentic AI application. AAL defines *what an agentic system is authorized to do in deployment*, rather than what it is technically capable of doing. An agent should not operate at an AAL higher than the level of accountability the human organization is prepared to accept. Evaluating AAL therefore requires examining the AI system within its full operating environment, including the LLM, the agent harness, and the surrounding organizational, procedural, and contextual constraints.

Table 1 summarizes the five levels of Allowed Autonomy. The progression from lower to higher AAL reflects increasing degrees of independent action, scope of impact, and transfer of control from human operators to AI systems. As autonomy increases, human control becomes less direct, reversibility decreases, and the consequences of failures become more difficult to contain, requiring correspondingly stronger governance justification and safeguards.

At **Level 1 (A1)**, agents execute bounded tasks only in response to explicit human invocation. Agents do not act autonomously and do not initiate actions on their own. Humans are fully hands-on and execute tasks by calling agents. The scope of agent operation is limited to individual, local tasks. If an agent makes a mistake, the consequences are bounded to a single invocation and can be corrected directly by the human who initiated the action.

Table 1: Allowed Autonomy Level (AAL).

| AAL | Human Role | Human Tasks | Task Scope | Accountability Implication |
|---|---|---|---|---|
| A1 | Hands on | Execute | Local-task | Errors are bounded to a single invocation and corrected directly by the human who initiated the action. |
| A2 | Instructions on | Instruct | Decision-support | The agent provides advice, which can indirectly degrade human decisions. |
| A3 | Eyes on | Supervise | Workflow | Agents now act; failures become actions rather than advice, though supervision and reversibility still limit harm. |
| A4 | Mind on | Set goals and constraints | System | Humans no longer supervise steps; failures stem from design errors rather than execution mistakes. |
| A5 | Mind off | Set policies | Organizational | Decision-making authority is delegated; failures reflect accountability and governance gaps, not task errors. |

At **Level 2 (A2)**, agents provide suggestions while being called or while interacting with humans, but do not take actions except generating recommendations, analyses, or proposed steps in response to human instructions or prompts, but all execution remains human-performed. The human's role is to issue instructions and make decisions based on the agent's output. The agent's scope is limited to decision support. Errors or flaws in agent outputs may influence human decisions, but the agent itself does not act on the environment.

At **Level 3 (A3)**, agents execute multi-step tasks under human supervision. Actions taken by agents require human approval, must be reversible to a previous step, or must be accompanied by mitigation mechanisms when actions are taken without explicit approval. At this level, agents operate at the workflow scope rather than at the individual task scope. Since agents now execute actions, failures manifest as unwanted actions rather than advice; however, supervision and reversibility help limit potential harm. Reversibility is highly application-dependent and should be carefully evaluated. For example, an AI coding agent that commits changes to a source-control repository can be considered reversible, whereas an agent that generates and executes control commands for an industrial chemical process is not.

At **Level 4 (A4)**, agents autonomously execute open-ended, multi-step workflows to achieve human-defined goals within fixed constraints. Humans specify goals, success criteria, and guardrails upfront, but do not approve individual actions during execution. Actions at this level may be consequential or irreversible, moving correctness concerns from runtime supervision to upfront design. Humans remain available for exception handling and escalation but do not oversee individual steps. The agent's scope expands from workflows to system-level operation, and failures at this level more often result from design or specification errors than from execution mistakes.

At **Level 5 (A5)**, humans define, encode, and authorize policy intent and operating boundaries before operation begins. Once authorized, agents operate autonomously within these policies, making and executing decisions without human intervention, step-level supervision, or routine reversibility. Accountability rests with the organization that delegates this authority. At this level, agents operate at an organizational scope, and failures reflect breakdowns in governance, policy design, or accountability rather than task-level execution errors.

Levels A3, A4, and A5 represent successive transfers of control from runtime supervision to upfront governance and, ultimately, to delegated authority. At A3, agents execute actions but remain supervised, interruptible, and reversible. At A4, humans no longer supervise execution and instead rely on upfront goal specification and constraints, even when actions may be irreversible. At A5, operational authority itself is delegated, with agents acting autonomously within authorized policies and boundaries.

To help development teams reason about these levels, we provide informal descriptors based on the human role involved: "**hands on**" (A1), "**instructions on**" (A2), "**eyes on**" (A3), "**mind on**" (A4), and "**mind off**" (A5). We expect most real-world Agentic AI applications to operate at A1 through A3. Levels A4 and A5 require significantly stronger justification and safeguards due to their scope and potential impact and are therefore expected to be assigned more selectively.

Given the wide range of AI applications within a large enterprise like us and the need for a framework that can be applied consistently across use cases, we found that five AAL levels strike an appropriate balance between distinguishing meaningful differences in autonomous behavior and maintaining usability for governance reviews. Organizations operating in different contexts may conclude that a different number of levels better suits their needs.

## Autonomous Capability Level

Autonomous Capability Level (ACL) concerns the *inherent technical capabilities* of an agent, independent of whether those capabilities are exercised in deployment. ACL characterizes what an agent *can do*, rather than what it is *authorized to do*, and focuses on the agent's functional abilities rather than organizational or governance considerations. In the autonomy decision framework introduced earlier, ACL assessment informs—but does not determine—the Allowed Autonomy Level. Table 2 summarizes the five Autonomous Capability Levels defined in this section.

Table 2: Autonomous Capability Level (ACL).

| ACL | Capability | Agent As Example Roles | Highest AAL | Example |
|---|---|---|---|---|
| C1 | High-level perception/low-level reasoning. Able to provide results when called. | Model, Tool | A1 | Pattern recognition model, isolated manufacturing systems |
| C2 | High-level reasoning. Able to help humans make decisions. | Helper, Advisor, Assistant | A2 | Forecasting, predictive maintenance |
| C3 | Reasoning, planning, and tool calling. Able to complete tasks defined by humans. | Junior Operator | A3 | AI data analysis, some AI coding tools. |
| C4 | Reasoning, multi-step planning, tool orchestration. Able to plan and complete tasks from a goal: execute open-ended workflows. | Sr. Operator | A4 | Advanced AI coding agents |
| C5 | Reasoning, planning, tool orchestration, monitoring, and adaption. Able to run entire operations independently. | Plant Manager, Decision Maker | A5 | Self-running supply chain, unmanned manufacturing |

We classify agent capabilities into five levels based on the progression of abilities in perception, reasoning, planning, action, and adaptation. At the lowest level, denoted as **C1**, an agent possesses high-level perception or low-level reasoning. For example, a computer vision model that converts raw camera input into labels of different objects is considered to exhibit high-level perception. It is important to note that perception and reasoning may overlap, and that it is often difficult to draw a strict boundary between high-level perception and low-level reasoning. For instance, in an autonomous driving scenario, after detecting an object, an agent may estimate the object's speed and trajectory and determine whether a collision is likely. This capability lies on a continuous spectrum, and the boundary between perception and reasoning is blurred. At this level, agents do not possess advanced reasoning, planning, or action capabilities and may take the form of models or tools. An agent at this level is a clear floor for agent capabilities, recognizing that many traditional machine learning applications continue to play an important role in modern AI systems and can be classified at this level.

At **C2**, an agent begins to demonstrate high-level reasoning and can assist humans in decision-making. As an example, agents at this level act as helpers, advisors, or assistants to humans. Additional examples include market forecasting agents and predictive maintenance agents. AI assistant tools that provide recommendations to humans can also be classified at this level. Agents at C2 may qualify as AI agents under our definition. Although the agents' planning and action capabilities remain limited. The examples provided here are not exhaustive, but rather illustrative within prescribed boundaries. For instance, if market forecasting is performed solely using a time-series model, the system should be classified as C1. If, however, the agent invokes a time-series model, incorporates market sentiment analysis, and reasons over internal business knowledge to forecast market trends, it should be classified as C2.

At **C3**, an agent is fully capable of perception, reasoning, planning, tool invocation, and taking actions. There is a significant increase in capability from C2 to C3, and agents at this level are unequivocally AI agents under our definition. Given these capabilities, C3 agents can be viewed as junior operators. Humans define tasks at a high level, such as "analyze this dataset." The agent may autonomously write code, execute the analysis, interpret the results, and visualize the outputs.

Agents at **C4** possess advanced reasoning, multi-step planning, and tool orchestration capabilities. They can plan and complete open-ended workflows based on goals specified by humans. Agents at this level may be considered senior operators. For example, instead of requesting "analyze this dataset," a human can ask an agent to "build a market analysis report." The C4 agent then searches for relevant data, performs analyses (like a C3 agent), generates charts, writes reports, revises content, and presents its results. Some of the most advanced AI-assisted coding tools operate at this level.

At **C5**, agents exhibit perception, reasoning, planning, tool orchestration, monitoring, and adaptation capabilities sufficient to operate independently over extended periods.

Agents, at this level, can be viewed as managers or decision-makers. Example applications include self-running supply chains and unmanned manufacturing systems. It remains debatable how many Agentic AI applications currently operate at this level. For example, some may consider a software system like OpenClaw (Wang et al. 2026) or similar Agentic AI systems to approach C5. Regardless, agents at this level are associated with high risk and potentially severe real-world consequences.

In our experience, five capability levels provide sufficient granularity to distinguish meaningful differences in agent capabilities while remaining practical for governance reviews. Other organizations may reasonably choose different level boundaries. For example, C1 and C2 could be combined if less distinction is needed among lower-capability systems. Conversely, organizations that place greater emphasis on learning, adaptation, or self-improvement capabilities may choose to introduce additional capability levels.

## Example Applications

In this section, we use two deployed applications to illustrate how Autonomous Capability Levels and Allowed Autonomy Levels can be applied to manage risk. Additional details are provided in the supplementary material.

The first system detects and remediates data quality gaps within a large-scale enterprise platform that integrates real-time drilling sensor data for operational surveillance. A supervisor agent routes natural language requests to specialized sub-agents for data gap detection, log analysis, and remediation execution. This multi-step goal decomposition is the characteristic of C4. The system is deployed at A3 rather than the technically feasible A4 because ungoverned backfill jobs can corrupt downstream pipelines.

The second system supports engineering decision-making in unconventional oil field operations, where engineers review ~20,000 unstructured field reports annually and incorrect conclusions can inform unsafe decisions about high-value assets and personnel safety. A supervisor routes each query to specialized sub-agents for text-to-SQL retrieval, RAG over operating procedures, vision-language analysis of engineering drawings and pump performance monitoring cards, deterministic failure signal extraction, time-series surveillance, and physics-based equipment sizing. Due to the risk of operational hazards, every reasoning step is constrained by deterministic verification tools and explicit guardrails, following a "tools verify, humans decide" operating principle. The system is technically capable of C3, yet deliberately deployed at A2 (advisory-only) because correct interpretation requires domain-specific engineering judgment, physical-domain actions are largely irreversible, and failures can carry significant safety and asset consequences.

Together, the two deployments demonstrate the framework's effectiveness in managing risk by decoupling agents' capabilities from permitted autonomy. While both applications assign an AAL that is one level below the corresponding capability upper bound, this is not always necessary. Deploying a system at its maximum allowable autonomy level may be appropriate when risks are well understood and adequately managed.

## Discussion and Future Work

As discussed earlier, this work builds on an established Responsible AI framework that has served effectively prior to the widespread adoption of Agentic AI applications. The underlying governance principles, intake processes, and organizational accountability structures remain sound and do not require fundamental revision. Instead, the primary gap exposed by Agentic AI lies in the absence of clear definitions for AI agents and Agentic AI, as well as the lack of a structured decision process for assigning appropriate autonomy levels.

The framework introduced in this paper addresses these gaps by explicitly separating capability assessment from authorization decisions and by providing a practical mechanism for integrating Agentic AI into existing governance processes. The framework is intentionally designed to be intuitive to promote adoption and usability. The terminology used to describe different levels—such as “junior operators” and “senior operators” for capability levels, and “hands on” and “mind on” for allowed autonomy levels—has proven effective for communication but may benefit from further fine-tuning as the framework is applied more widely in practice.

The five-level structures for both AAL and ACL have proven practical within our organization, balancing expressive power with governance usability. Other organizations may adopt more or fewer levels, provided the underlying separation between capability and authorization is preserved. The level boundaries, especially for AAL, may be refined based on different organizations’ needs. Readers may refer to seminal work by Sheridan and Verplank (1978) to gain insight into such refinement.

The design outlined here is to refine our existing Responsible AI framework. As more applications go through this process, practical lessons will emerge regarding usability, calibration, and consistency across teams and application domains. Future work will focus on evaluating the framework across these applications, refining autonomy level definitions and terminology where needed, examining how level-assignment judgments vary across applications, and sharing empirical learnings from deployment and governance practice in future venues.

# Appendix

## A. Data Engineering Multi-Agent System

### A.1 System Architecture

The system is designed to detect and remediate data quality gaps within a large-scale enterprise data platform. It employs a supervisor-routed multi-agent architecture: a central orchestrator receives user requests and routes them to specialized sub-agents responsible for bounded domains. Each sub-agent's reasoning chain is deliberately kept short, typically three to four steps, reflecting current limitations in LLM reliability over longer chains. As shown in Figure A1, the supervisor classifies user intent and delegates tasks to four specialized sub-agents: Data Quality, Job Log Analysis, Real-time Data Gap Detection, and Azure Data Explorer (ADX) Gap Detection. Each equipped with domain-specific tools and remediation capabilities. This intent-based routing logic decomposes natural language requests into actionable workflows, enabling autonomous data pipeline monitoring, root cause analysis, and automated gap remediation across the Azure Databricks and Azure Data Explorer ecosystem.

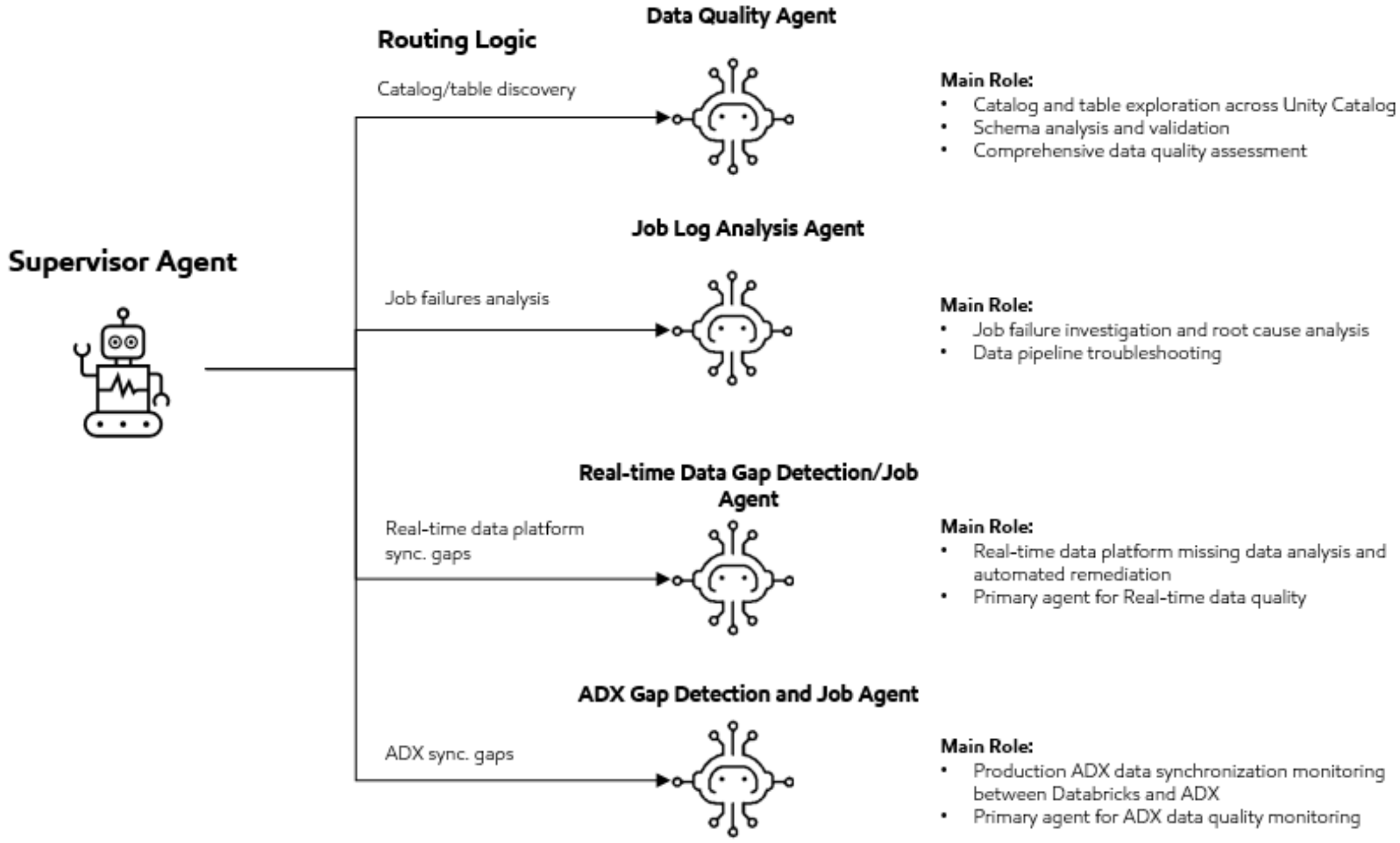

Figure A1: Data Engineering Multi-Agent System.

## A.2 Capability Assessment (C4)

From a capability perspective, the composed system exhibits characteristics consistent with C4 (Senior Operator). While each individual sub-agent operates at C3 (autonomously querying metadata catalogs, comparing data states against ground truth, and invoking tools), the system demonstrates multi-step planning and tool orchestration working towards a high-level goal. A user request to "find and fix data gaps" results in the supervisor routing to a gap detection sub-agent, then invoking a log analysis sub-agent to diagnose root causes and finally triggering a remediation job through a job execution sub-agent. The behavior of decomposing a goal into sub-tasks, orchestrating multiple specialized agents, and synthesizing their outputs, is the distinguishing characteristic of C4. This assessment highlights an important consideration: in multi-agent systems, the Autonomous Capability Level should be evaluated at the level of the composed system, not at the level of individual components in isolation. This is consistent with our definition of Agentic AI, where autonomous behavior emerges from interactions between multiple components rather than from any single agent. The system does not reach C5 because it lacks continuous self-monitoring and adaptation capabilities, and reasoning chains are deliberately bounded.

## A.3 Deployment Decision (A3)

Although the agent is capable of C4: goal-directed, multi-step workflow execution, the system is deployed at A3 (Eyes On) rather than A4. Certain remediation actions, such as triggering backfill jobs, carry significant operational risk if executed without oversight. An un-governed backfill job can corrupt downstream data pipelines, introduce silent data-quality regressions, or consume substantial compute resources unnecessarily. This risk profile motivates restricting the system to A3, where a human supervisor monitors outcomes, approves consequential actions, and retains the ability to intervene or reverse them. This deployment decision is justified by the domain's favorable risk characteristics: outputs are independently verifiable against ground truth, reasoning chains within each sub-agent are short and decomposable, and remediation actions are reversible. Architectural controls, including identity propagation from user through supervisor to tool, programmatic loop-count guardrails, and immutable audit traces, provide the operational safeguards necessary to sustain supervised autonomy while preventing unsanctioned escalation to A4.

An alternative deployment at A2 (Instructions On) is also viable for the same system. In a C4/A2 configuration, the system identifies gaps and recommends remediation actions, but all execution remains human-performed. The framework accommodates both configurations without re-assessing the underlying capability level, and an organization may transition between A2 and A3 as infrastructure matures and operational confidence grows. The system does not meet the conditions for A4 (Mind On), which would require unsupervised execution of open-ended workflows. This would be a level of delegation the organization is not prepared to accept for a production-critical data platform, regardless of the agent's technical capabilities. The gap between C4 and A2 or A3 is not a deficiency; it is a deliberate governance choice. The framework supports this by design: an organization can deploy at C4/A2, build confidence, mature its rollback infrastructure, and advance to C4/A3, or revert from A3 to A2 if operational conditions change, without re-assessing the underlying capability level.

## B. Energy Operations Multi-Agent Copilot

### B.1 System Architecture

The system is designed to accelerate engineering decision-making in unconventional oil and gas field operations, where engineers must review approximately 20,000 unstructured field reports annually, including free-text job logs, inspection records, and equipment drawings. As shown in Figure A2, it employs a supervisor-routed multi-agent architecture in which a central orchestrator receives engineer queries and delegates tasks to specialized sub-agents responsible for bounded analytical domains: text-to-SQL retrieval against operational databases, retrieval-augmented generation over operating procedures and case files, vision-language analysis of engineering drawings and equipment images, deterministic failure-signal extraction with taxonomy validation and physics checks, time-series surveillance from sensor data, and physics-based equipment sizing including numerical simulations for tubing sizing and gas lift design. Each sub-agent uses short reasoning chains bound by deterministic verification tools under an explicit "AI reasons, guardrails verify, human decides" operating principle, because agent hallucination in a regulated and high-stakes industry is an operational hazard.

### B.2 Capability Assessment (C3)

From a capability perspective, the composed system is assessed at C3 (Junior Operator), with emerging characteristics of C4. The system reasons, plans, and calls tools to complete human-defined analytical tasks: autonomously formulates and executes database queries, invokes computational sizing routines, and synthesizes extracted signals into structured findings. The system's autonomous query formulation and tool execution are hallmarks of C3. The system stops short of C4 considering its supervisor follows an established workflow and performs intent-based routing to the appropriate specialist for a given query rather than autonomously

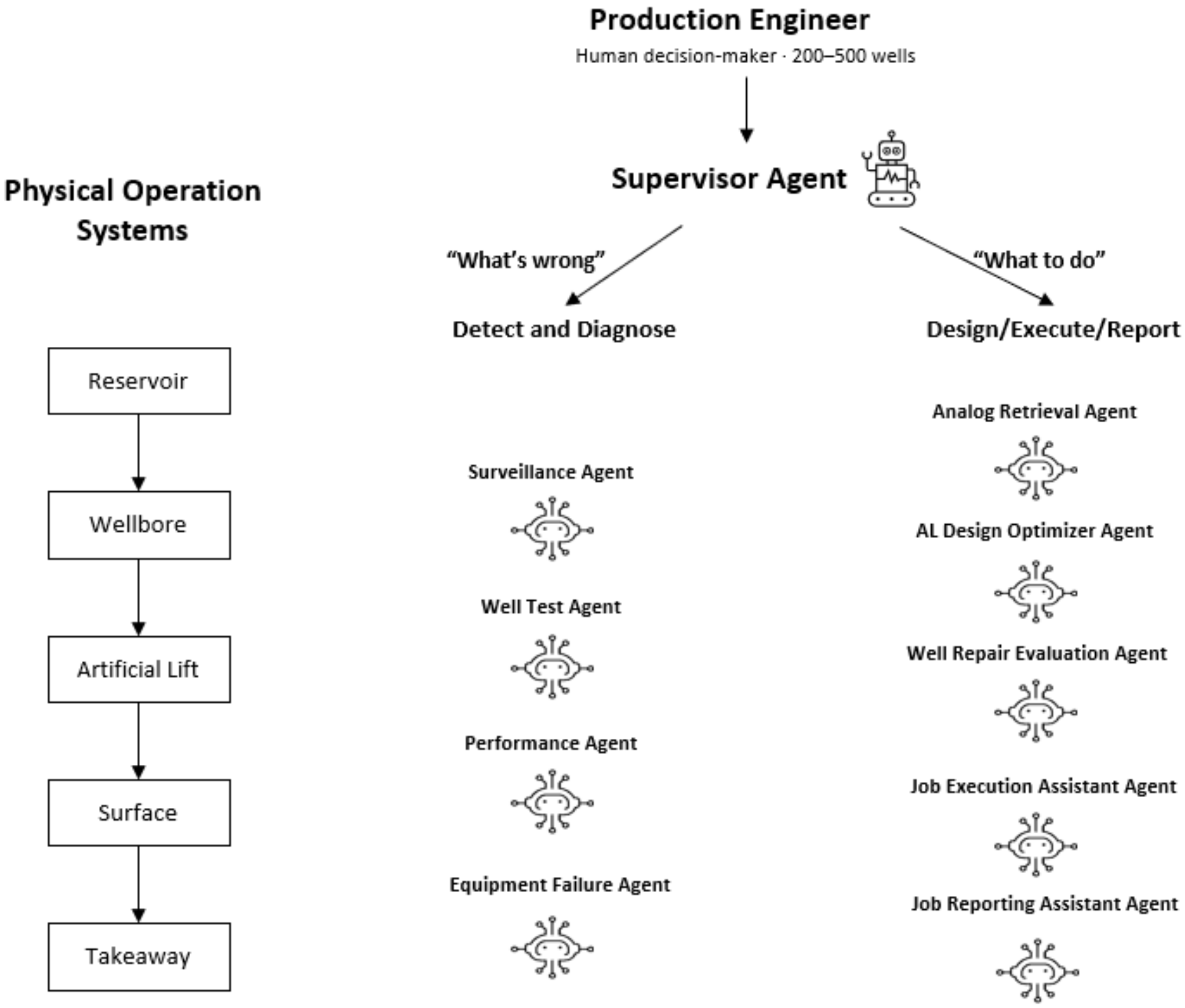

Figure A2: Energy Operations Multi-Agent Copilot Architecture.

.

decomposing an open-ended goal into a multi-agent execution plan, and it does not exhibit the continuous self-monitoring and adaptation that would indicate C5.

The contrast with the C4 data engineering system is instructive: the data engineering agent chains gap detection, root-cause diagnosis, and remediation into a pipeline pursuing a compound goal ("find and fix"), whereas the energy operations copilot dispatches each bounded query to a single specialist.

## B.3 Deployment Decision (A2)

Despite being technically capable of operating at A3, the system is purposefully deployed at A2 (Instructions On). It analyzes reports, answers operational questions, and surfaces recommended findings. All consequential decisions and actions remain human-performed. The agent does not execute actions against physical or production systems. This deployment decision is justified by the domain's unfavorable risk characteristics: outputs cannot be independently verified against external ground truth, and actions in the physical domain are largely irreversible (a wrong equipment sizing decision cannot be undone once installed downhole). At the meanwhile correct interpretation frequently depends on engineering judgment and tacit knowledge that engineering domain expertise must validate. Under these conditions, a failure at A2 manifests as a flawed recommendation that a human reviewer can catch, rather than as an unsanctioned action, keeping accountability squarely with the human decision-maker.

The system is nonetheless architecturally designed for incremental autonomy advancement. As deterministic verification matures for specific bounded, reversible sub-tasks, individual workflows may advance toward A3 without reassessing the underlying C3 capability